\documentclass{article}

\pdfoutput=1

\usepackage[final,nonatbib]{neurips_2022_ml4ps}

\usepackage[utf8]{inputenc} 
\usepackage[T1]{fontenc}    
\usepackage{hyperref}       
\usepackage{url}            
\usepackage{booktabs}       
\usepackage{amsfonts}       
\usepackage{nicefrac}       
\usepackage{microtype}      
\usepackage{xcolor}   
\usepackage{graphicx}
\usepackage{subcaption}
\usepackage{adjustbox}
\usepackage[normalem]{ulem}

\title{On Using Deep Learning Proxies as Forward Models in Optimization Problems}

\author{%
  Fatima Albreiki\thanks{Work done as part of the internship at AIQ.}, Nidhal Belayouni and Deepak K. Gupta \\
  AIQ, UAE \\
  \texttt{fatima.albreiki@mbzuai.ac.ae, \{nbelayouni,dgupta\}@aiqintelligence.ae} \\
}

\begin{document}

\maketitle

\begin{abstract}
Physics-based optimization problems are generally very time-consuming, especially due to the computational complexity associated with the forward model. Recent works have demonstrated that physics-modelling can be approximated with neural networks. However, there is always a certain degree of error associated with this learning, and we study this aspect in this paper. We demonstrate through experiments on popular mathematical benchmarks, that neural network approximations (NN-proxies) of such functions when plugged into the optimization framework, can lead to erroneous results. In particular, we study the behaviour of particle swarm optimization and genetic algorithm methods and analyze their stability when coupled with NN-proxies. The correctness of the approximate model depends on the extent of sampling conducted in the parameter space, and through numerical experiments, we demonstrate that caution needs to be taken when constructing this landscape with neural networks. Further, the NN-proxies are hard to train for higher dimensional functions, and we present our insights for 4D and 10D problems. The error is higher for such cases, and we demonstrate that it is sensitive to the choice of the sampling scheme used to build the NN-proxy. The code is available at \url{https://github.com/Fa-ti-ma/NN-proxy-in-optimization}. 

\end{abstract}

\section{Introduction}
In the past few years, deep learning has led to several advancements in the different disciplines of science and engineering. Some popular examples include the development of a rigorous solution to the protein-fold problem \cite{jumper2021alpha} and enhancing computational fluid dynamics simulations \cite{vinuesa2022cfd}, among others. One popular application of deep learning is to develop proxy models that can serve as an alternative for the computationally-intensive physics-based models. An interesting example is the deep-learning tomography method \cite{arayapolo2018tle} which aims at by-passing the computationally demanding steps of the traditional seismic tomography processes.

The gain with deep learning can be even higher when such a proxy model is used as an alternative to a physics-based simulation that serves as a forward model of an optimization problem. Note that the optimization problem referred is different from the optimization problem solved for model training in deep learning. We refer here to the physics-based optimization problems where the goal is to optimize the parameters associated with the underlying physics of a certain problem. An example is the problem of near-surface velocity modeling of the earth, where the goal of the optimization is to identify the distribution of compression and shear velocities for a predefined model space, such that the resultant spectrum images of the earth, as acquired by sensors, can be optimally \mbox{reconstructed \cite{eage2022zwartjes}}. This optimization involves solving the forward model and computing its gradient repeatedly for a number of iterations, and reducing the cost associated with the forward model can speed up the overall optimization problem by a large margin. 

When using deep learning proxy models, it is important to note that the parameter space of the proxy model might not be the same as the original forward model. While the physics-model is parameterized with a few control variables, the proxy NN-model is generally designed using millions of weights. The choice of parameters for the NN-proxy depends on several factors, such as the sampling method for picking the training data, extent of convergence of the model, \emph{etc}. It is well known that the performance of the optimization methods depends significantly on the complexity of the parameter landscape, and a changed landscape due to the use of deep learning can also lead to completely different optimized solutions which are not necessarily optimal. Clearly, there are two different aspects that need to be studied when creating such a proxy model for the optimization process: the robustness of the deep learning approach as well as efficacy of the chosen optimization method. 

\textbf{Contributions. }In this work, we present our first observations on how the choice of data sampling as well as the amount of data samples for training the NN-proxy can influence the performance of the overall optimization process. We use the popular benchmark functions from the field of global optimization (\emph{Rosenbrock, Rastrigin, Ackley, \emph{etc.}}) as alternative representations for the physics models used in NN-proxies. These mathematical functions have long served as benchmarks for the generic field of optimization, and the exact solutions are known. For optimization, we use particle swarm optimization (PSO) \cite{PSO} method and genetic algorithm (GA) \cite{GA} and start with first analyzing the stability of these methods when the original parameter landscapes are replaced with those obtained from NN-proxies. Our investigation reveals that the chosen optimization methods become very sensitive to their respective initialization schemes when the NN-proxy is employed. Sampling scheme as well as the amount of training data plays an important role in the construction of good-NN proxies, and we study this aspect as well qualitative as well as quantitative assessment of the error in optimization. Lastly, we also study the influence of NN-proxy error on optimization in higher dimensions and report results for 4D and 10D functions.

\section{Deep Learning Proxy Models (NN-Proxies)}
\begin{figure}
    \centering
    \includegraphics[scale=0.33]{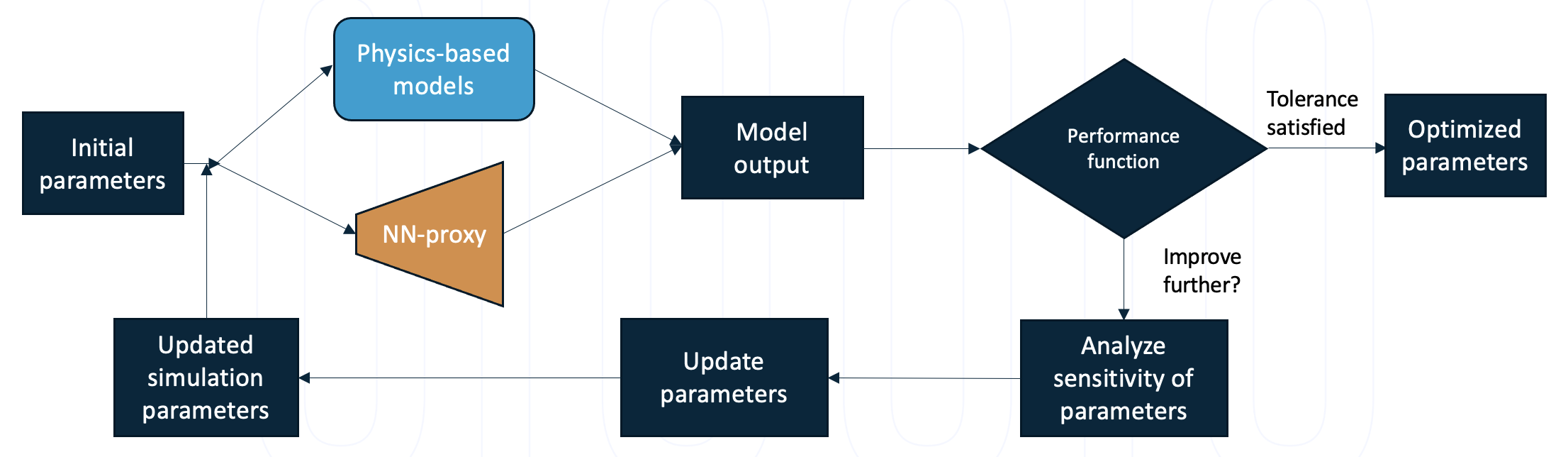}
    \caption{Schematic representation of a conventional optimization pipeline using conventional physics-based forward models as well as their alternative representation using NN-proxies.}
    \label{fig:schema}
\end{figure}

Building a deep learning proxy model, also referred as NN-proxy, implies learning the mapping of the physics-based models in a data-driven manner and using it as a forward model in an optimization problem. A better understanding of this can be obtained from Figure \ref{fig:schema}. Similar to the physics-model, the NN-proxy computes the desired output which is then evaluated using the objective function to decide if the model input parameters are to be updated or not. It is evident from the figure that a large error in the predictions made by the NN-proxy can drift the whole optimization process off track, and it is very important that the neural network model is sufficiently good before it can be plugged into the optimization process as a NN-proxy. We analyze this aspect through multiple experiments on popular mathematical benchmarks below.

\textbf{Experimental setup. }There are several factors that need to be understood when using a NN-proxy in a optimization process, and in this paper, we study a few important ones. In place of the original physics-models, we use popular mathematical benchmarks to study the efficacy of global optimization methods, and use neural networks to learn proxy representations of these functions. We use Rosenbrock, Rastrigin and Ackley functions \cite{rosen,rastrigin1974systems,d.h.ackley1987a-connectionist} and experiment with particle swarm optimization (PSO) \cite{PSO} and genetic algorithm (GA) \cite{GA} as the two optimization methods. The two chosen optimization methods are well established to solve complex multimodal global optimization methods, and we experiment here how they fair when the original models are replaced by their NN-proxies. We also consider how the error associated with the NN-proxy scales in higher dimensions, and for this purpose, we consider the cases of 4D and 10D. More details related to the chosen benchmark functions are described in Appendix \ref{sec-app-benchmarks}. Further, brief descriptions of the chosen optimization methods are presented in Appendix \ref{sec-app-optmethods}.

Below we present our observations related to a series of experiments aimed at understanding the efficacy of NN-proxies.

\begin{figure}
\centering 
\fbox{\includegraphics[scale=0.254]{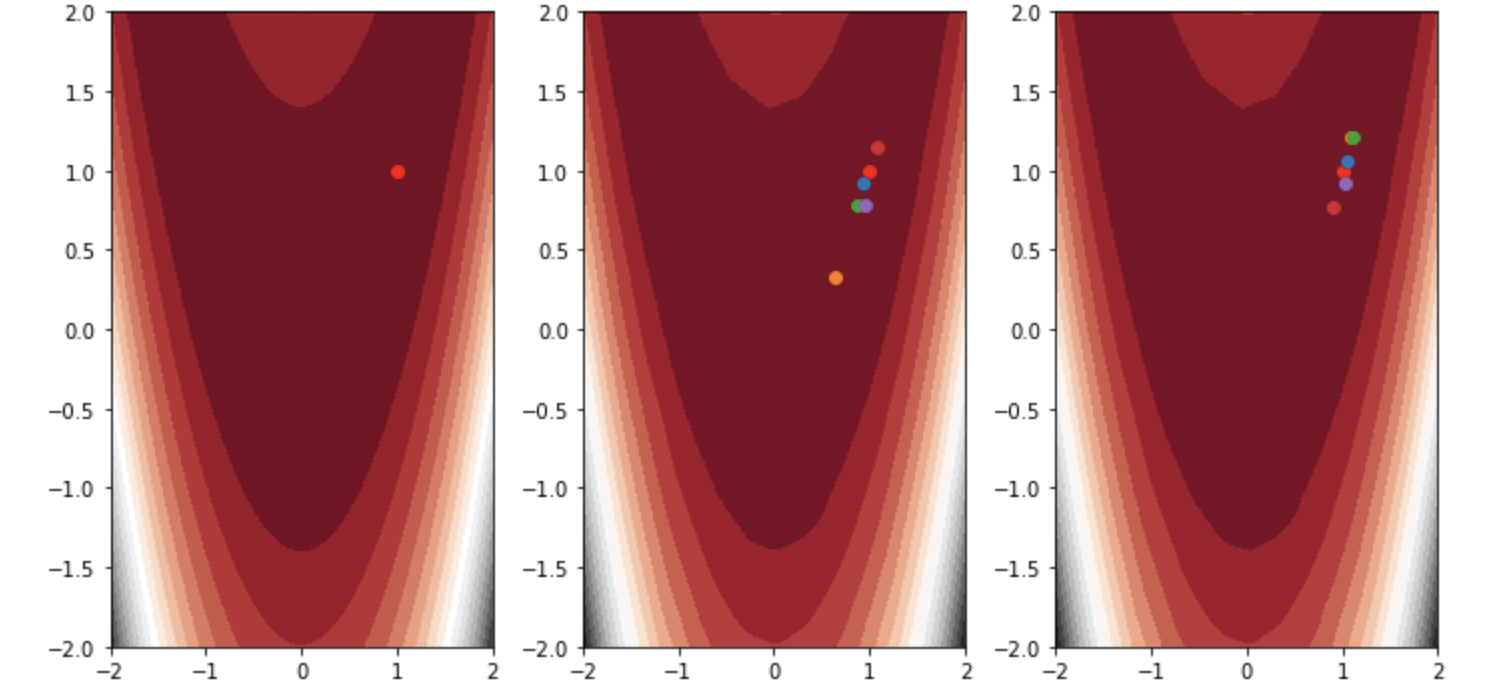}}
\fbox{
\includegraphics[scale=0.25]{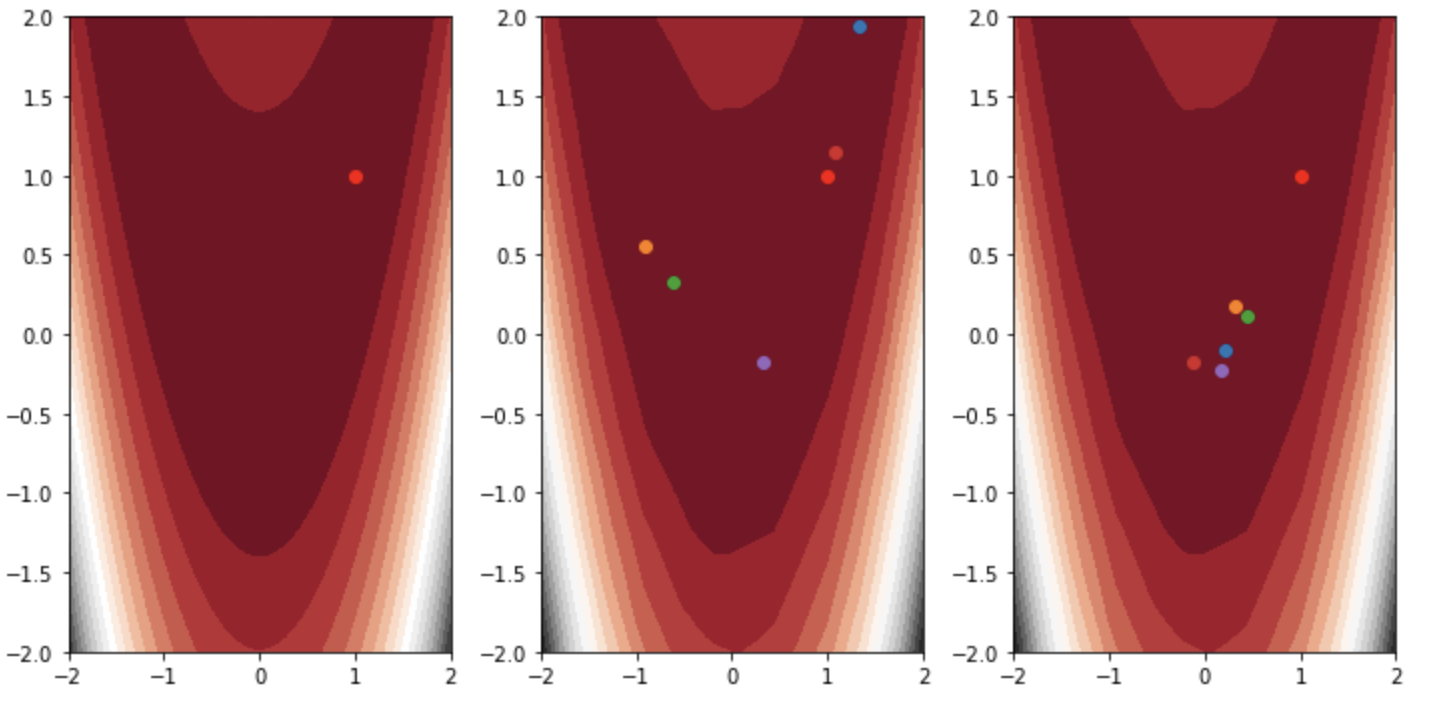}}
\caption{Example results for the Rosenbrock function for dense (left) and sparse (right) sampling of the function space. For each case, we show PSO and GA results on the true function (left), results of PSO on the NN-proxy (middle) and results of GA o the NN-proxy (right) for 5 random seeds.}
\label{fig:rosenbrock_seeds}
\end{figure}

\begin{figure}
\centering 
\fbox{\includegraphics[scale=0.105]{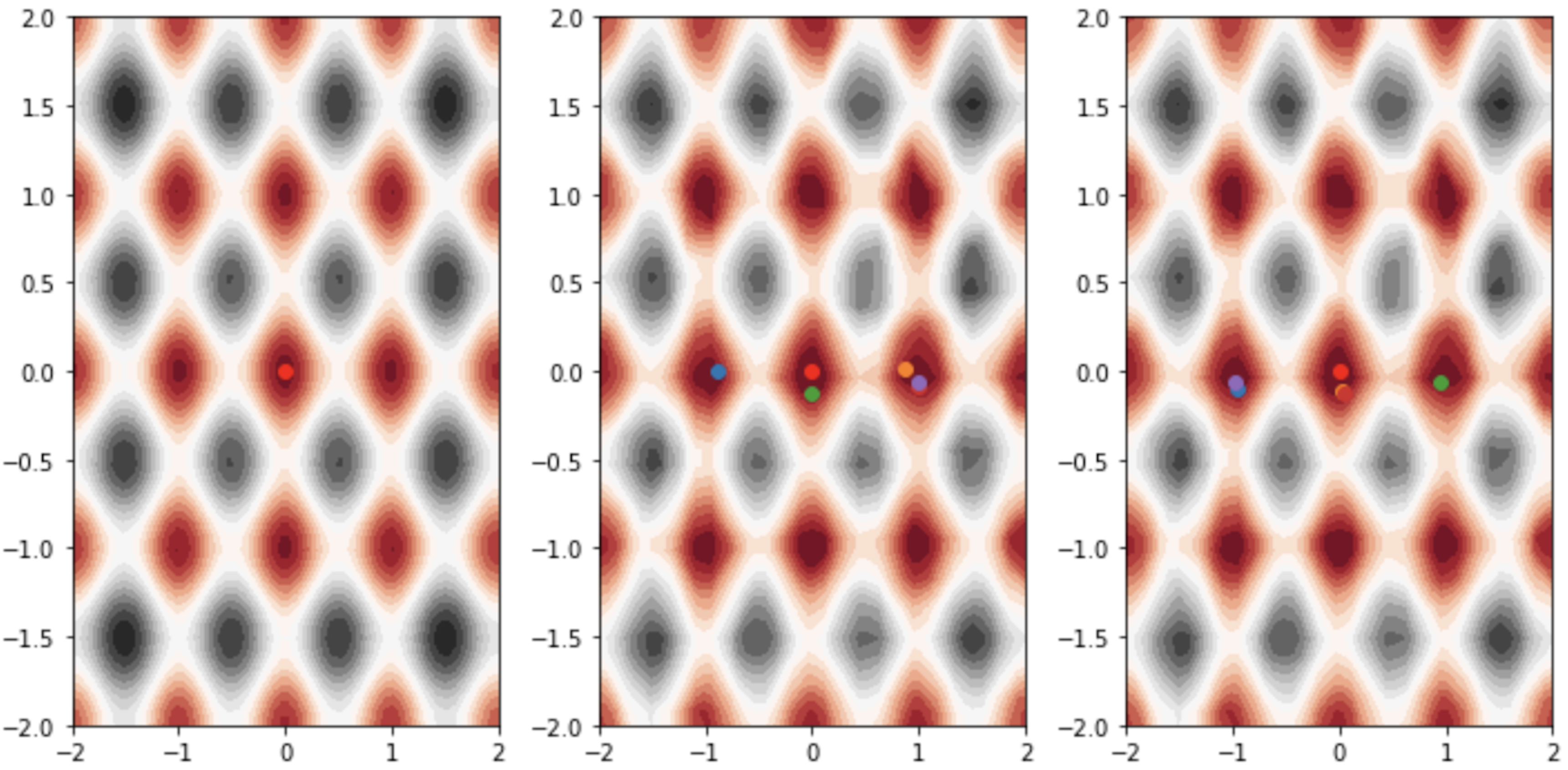}}
\fbox{
\includegraphics[scale=0.105]{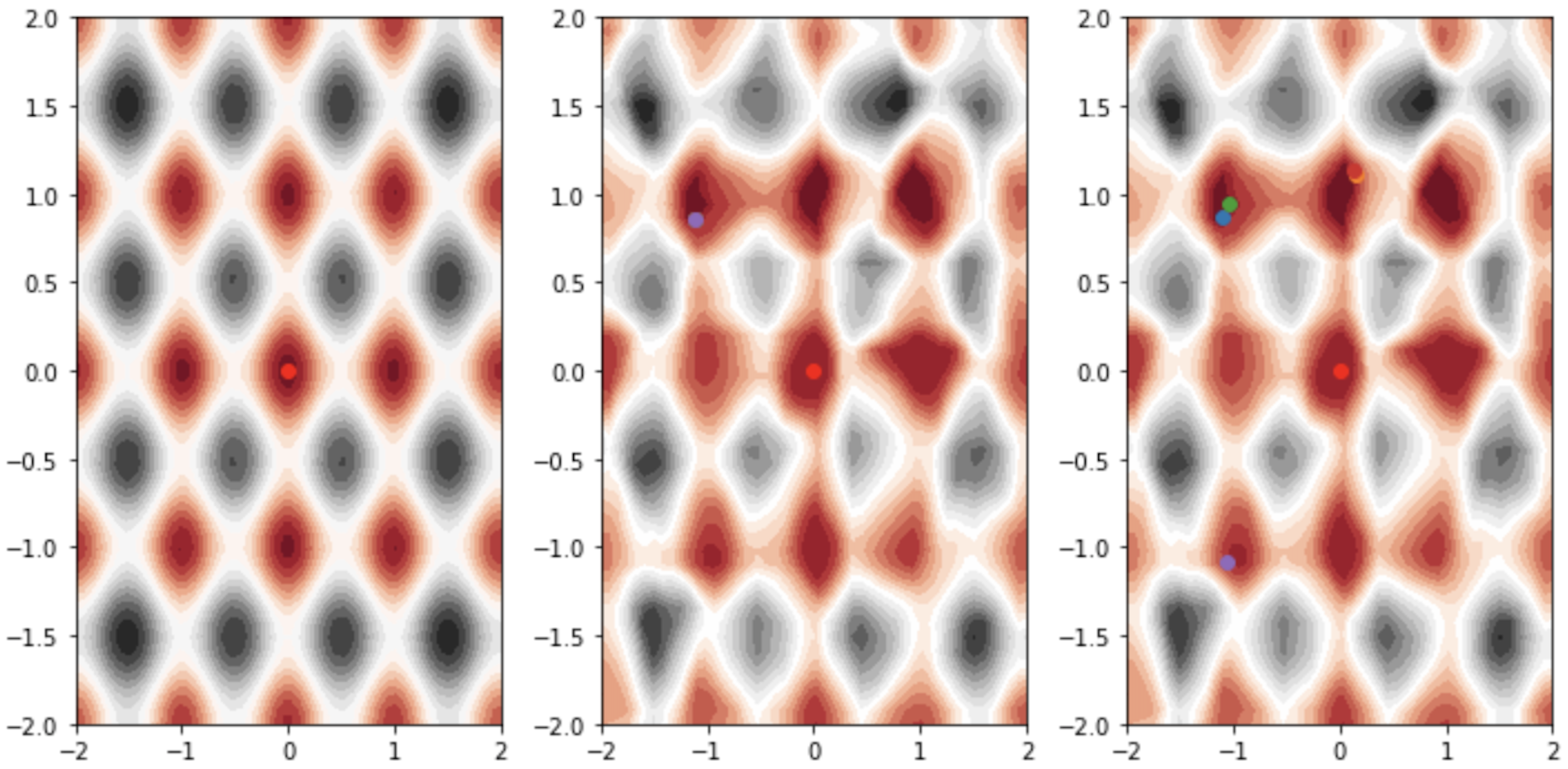}}
\caption{Example results for the Rastrigin function for dense (left) and sparse (right) sampling of the function space. For each case, we show PSO and GA results on the true function (left), results of PSO on the NN-proxy (middle) and results of GA o the NN-proxy (right) for 5 random seeds.}
\label{fig:rastrigin_seeds}
\end{figure}

\textbf{Choice of the initial model for optimization. }Optimization methods are generally sensitive to the choice of the initial values of the optimization parameters. Thus, it is of interest to investigate how this behaviour changes when the NN-proxies replace the actual forward models. Figures \ref{fig:rosenbrock_seeds} and \ref{fig:rastrigin_seeds} show the results obtained for Rosenbrock and Rstrigin functions, respectively, using PSO and GA. For both the functions, we present the solution of PSO and GA on the actual function, and it is observed that for almost all cases, the global minima is found. On the contrary, when the true function is replaced by its NN-proxy, both the optimization methods become very sensitive. For the case where dense sampling is used for creating the training dataset, the sensitivity to the initial values is low. However, for sparse sampling, the error is significantly high with the predicted solutions diverging significantly from the global minimum. Experimental details are described in Appendix \ref{sec-app-exp} and the quantitative results related to this set of experiments as well as for Ackley function are presented in Table \ref{tab:E2}. From this experiment, it can be concluded that the behavior of an optimizer with the actual physics model can be very different from that when used with a NN-proxy. Moreover, most physics problems are complex and computationally expensive, and dense sampling is not the choice, and with sparse sampling, the results can be very erroneous. This is a matter of concern and deserves attention when using NN-proxies to represent physics of a system.


\textbf{Effect of data sampling. }Data sampling plays an important role in bulding a good NN-proxy. This is clearly reflected from the results shown in Figures \ref{fig:rosenbrock_seeds} and \ref{fig:rastrigin_seeds} for 2D functions. We further study this and present a quantitative analysis in Table \ref{tab:E2}. Apart from the uniform dense and sparse samplings, we also include Gaussian sampling around the global minimum. This choice is meant to weakly reflect the scenario when a priori information is available on the location of the best solution. From the obtained results, we see that for all cases of sampling, there is a significant amount of error in the prediction when using the NN-proxy. In general, we see that sparse sampling is not very effective and the optimizers are very sensitive to the landscape constructed from such data. For Rosenbrock, GA seems to be unstable even for the true function. In terms of comparing the two optimization methods, both seem to work better than each other for different scenarios and equally good overall. Out of the three functions, we see that the results of PSO and GA are very good on the NN-proxy of Ackley function. The reason could be that the globally optimal solution of this function differs significantly from the locally optimal solutions and the NN-proxy can represent it very well.

\begin{table}
\caption{Mean Euclidean distance and the standard deviation between the globally optimal point of the three mathematical benchmarks in 2D and the respective solutions obtained for the NN-proxies using PSO and GA. Three different sampling methods are used, and ground-truth refers to the case where PSO and GA are used on the true functions.}
\begin{adjustbox}{width=1\textwidth}
\begin{tabular}{l|l|l|l|l|l|l}
Function  & \multicolumn{2}{c|}{Rosenbrock} &  \multicolumn{2}{c|}{Rastrigin} & \multicolumn{2}{c}{Ackley} \\ \hline
Optimizer & PS         & GA    & PS  & GA  & PS & GA \\ \hline
Ground-Truth     &    $0.0\pm 0.0$  &  $0.91\pm 0.53$  &  $0.0\pm 0.0$  &  $0.0\pm 0.0$   & 0.0  $0.0\pm 0.0$   &  $0.0\pm 0.0$  \\ \hline

Dense     &     $0.34\pm 0.21$      &   $0.54 \pm 0.69$    &  $0.75 \pm 0.42$ & $0.16\pm 0.22$   & $0.03 \pm 0.0$   &  $0.03 \pm 0.0$  \\ \hline
Sparse    &   $1.17 \pm 0.62$      &   $1.37 \pm 0.49$    &   $1.41 \pm 0.04$ &   $1.24 \pm 0.13$   &  $0.14 \pm 0.08$  &  $0.14 \pm 0.05$  \\ \hline
Gaussian     &      $1.07 \pm 0.54$      &    $1.15 \pm 0.28$   &  $1.07\pm 0.54$  & $1.15 \pm 0.28$    & $0.06 \pm 0.03$   &  $0.07 \pm 0.02$
\end{tabular} 
\end{adjustbox}
\label{tab:E2}
\vspace{-0.5em}
\end{table}

\begin{table}
\caption{Mean Euclidean distance and the standard deviation between the globally optimal point of the three mathematical benchmarks in 4D and the respective solutions obtained for the NN-proxies using PSO and GA. Here, three different sampling methods are used t build the training set for the NN-proxies.}
\begin{adjustbox}{width=1\textwidth}
\begin{tabular}{l|l|l|l|l|l|l}
Function  & \multicolumn{2}{c|}{Rosenbrock} &  \multicolumn{2}{c|}{Rastrigin} & \multicolumn{2}{c}{Ackley} \\ \hline
Optimizer & PS         & GA    & PS  & GA  & PS & GA \\ \hline
Dense     &    $0.60 \pm 0.03$       &   $1.59 \pm 0.74$    &   $0.47 \pm 0.0$  & $0.44 \pm 0.02$    &  $0.76 \pm 0.20$  & $0.90 \pm 0.20$   \\ \hline
Sparse    &     $0.86 \pm 0.06$       &   $1.56 \pm 0.72$    &   $0.72 \pm 0.0$  &   $0.70 \pm 0.02$   &  $0.37 \pm 0.03$  &   $0.27 \pm 0.07$ \\ \hline
Gaussian     &     $2.51 \pm 0.57$      &    $2.49\pm 0.85$   &  $0.71\pm 0.0$  & $0.67 \pm 0.05$   & $0.26 \pm 0.07$    &   $0.27 \pm 0.06$
\end{tabular} 
\end{adjustbox}
\label{tab:E4}
\vspace{-0.5em}
\end{table}

\begin{table}
\caption{Mean Euclidean distance and standard deviation between the true optimal solutions of the three mathematical benchmarks in 10D and the respective solutions obtained for the NN-proxies using PSO and GA.}
\begin{adjustbox}{width=1\textwidth}
\begin{tabular}{l|l|l|l|l|l|l}
Function  & \multicolumn{2}{c|}{Rosenbrock} &  \multicolumn{2}{c|}{Rastrigin} & \multicolumn{2}{c}{Ackley} \\ \hline
Optimizer & PS         & GA    & PS  & GA  & PS & GA \\ \hline
Dense     &     $2.62\pm 0.46$       &   $2.95 \pm 0.51$   &   $3.69 \pm 0.53$  & $4.45\pm1.06$    & $0.64 \pm 0.01$  & $0.70 \pm 0.13$ \\ \hline
Sparse    &     $3.37 \pm 0.74$      &  $3.71 \pm 0.51$     &   $2.20\pm 0.29$  &   $2.36\pm 0.56$  & $1.19 \pm 0.0$  & $1.05 \pm 0.08$  \\ \hline
Gaussian     &     $5.31 \pm 1.22$       &   $5.67 \pm 1.53$    &  $6.25 \pm 0.43$  &  $7.9 \pm 2.6$  & $0.30 \pm 0.0$    & $0.30 \pm 0.07$  
\end{tabular} 
\end{adjustbox}
\label{tab:E10}
\end{table}


\textbf{Stability in higher dimensions. }For a better approximation of the actual physics, the mathematical benchmarks need to be studied in higher dimensions. In this regard, we study the performance of NN-proxy in 4D and 10D also, and the related results are reported in Tables \ref{tab:E4} and \ref{tab:E10}, respectively. With the increased number of dimensions, the error grows as expected. An interesting observation is that for higher dimensions, the Gaussian sampling scheme, which is expected to better model the landscape arond the true solution, seems to perform inferior compared to sparse sampling. The anticipated reason behind this issue is that too many samples around the true solution limit the number of samples that can be used to approximate the landscape far from it. This means that the function cannot be constructed well for regions far from it. This can lead to landscapes comprising false good solutions for the NN-proxy. In this regard, a better choice seems to be to go with uniform sampling.


\section{Conclusions}
In this paper, we have studied how the outcome of a physics-based optimization process can be adversely influenced when approximating the underlying physics process with a neural network proxy. We have demonstrated through experiments on popular mathematical benchmarks, that neural network approximations (NN-proxies) of such functions can lead to erroneous results for different training setups. The correctness of the approximate model depends on the extent of sampling conducted in the parameter space, and caution needs to be taken when constructing this landscape with neural networks. Further, the NN-proxies are hard to train for higher dimensional functions, and through experiments on 4D and 10D problems, we demonstrated that the error in the optimized solution can be very high. The results reported in this paper are our first observations on the effect of NN-proxies on the optimization process. Clearly, we need to develop a rigorous benchmarking criterion to evaluate any novel NN-proxy and understand its robustness and generalization before it can be plugged in an optimization framework to replace a certain physics process.

\section{Broader Impact Statement}
This work focuses on understanding the error associated with the use of deep learning to replace the complex physics-based forward models used in optimization processes. While the initial concept in the paper is demonstrated on mathematical benchmarks, we hope that the observations and any countermeasures will be transferable on real world optimization problems involving physical simulations. Example of such problems includes proxy model for eigen value analysis in a compliant mechanism problem of topology and design optimization, among others. Overall, there are several problems where the results of this paper and the follow up will be useful.

Further, we do not see any ethical concerns of negative societal impact of this work.

\bibliography{neurips_2022}

\begin{thebibliography}{10}

\bibitem{d.h.ackley1987a-connectionist}
D.H Ackley.
\newblock {\em A Connectionist Machine for Genetic Hill climbing}, volume
  SECS28 of {\em The Kluwer International Series in Engineering and Computer
  Science.}
\newblock Kluwer Academic Publishers, Boston, 1987.

\bibitem{arayapolo2018tle}
Mauricio Araya-Polo, Joseph Jennings, Amir Adler, and Taylor Dahlke.
\newblock Deep-learning tomography.
\newblock {\em The Leading Edge}, 37(1):58--66, 2018.

\bibitem{deb2002ga}
K.~Deb, A.~Pratap, S.~Agarwal, and T.~Meyarivan.
\newblock A fast and elitist multiobjective genetic algorithm: Nsga-ii.
\newblock {\em IEEE Transactions on Evolutionary Computation}, 6(2):182--197,
  2002.

\bibitem{GA}
John~H. Holland.
\newblock Genetic algorithms.
\newblock {\em Scientific American}, 267(1):66--73, 1992.

\bibitem{jumper2021alpha}
J.~Jumper, R.~Evans, and A.~et~al. Pritzel.
\newblock Highly accurate protein structure prediction with alphafold.
\newblock {\em Nature}, 596:583--589, 2021.

\bibitem{PSO}
J.~Kennedy and R.~Eberhart.
\newblock Particle swarm optimization.
\newblock In {\em Proceedings of ICNN'95 - International Conference on Neural
  Networks}, volume~4, pages 1942--1948, 1995.

\bibitem{rastrigin1974systems}
Leonard~Andreevi{\v{c}} Rastrigin.
\newblock Systems of extremal control.
\newblock {\em Nauka}, 1974.

\bibitem{rosen}
H.~H. Rosenbrock.
\newblock {An Automatic Method for Finding the Greatest or Least Value of a
  Function}.
\newblock {\em The Computer Journal}, 3(3):175--184, 1960.

\bibitem{vinuesa2022cfd}
R.~Vinuesa and S.~L. Brunton.
\newblock Enhancing computational fluid dynamics with machine learning.
\newblock {\em Nature Computational Science}, 2:358--366, 2022.

\bibitem{eage2022zwartjes}
P.~Zwartjes, D.~Gupta, and T.~Gupta.
\newblock Near surface velocity estimation using surface waves and deep
  learning.
\newblock {\em Proceedings, EAGE}, pages 1--5, 2022.

\end{thebibliography}
\bibliographystyle{plain}

\appendix

\section{Benchmark Functions}
\label{sec-app-benchmarks}
\begin{figure*}
\begin{subfigure}{0.33\textwidth}
\centering
    \includegraphics[scale=0.27]{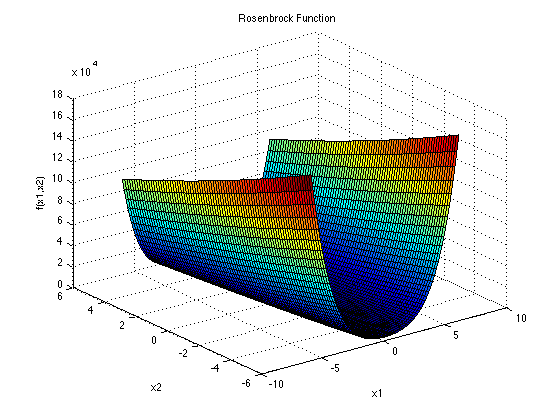} 
    \subcaption{Rosenbrock}
     \label{fig:rosenbrock}
\end{subfigure}
\begin{subfigure}{0.33\textwidth}
\centering
    \includegraphics[scale=0.25]{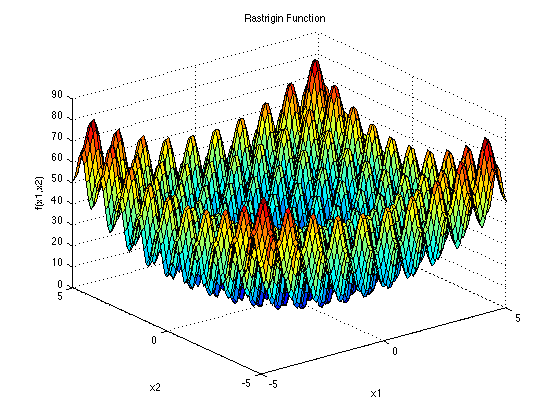} 
    \subcaption{Rastrigin}
     \label{fig:rosenbrock}
\end{subfigure}
\begin{subfigure}{0.33\textwidth}
\centering
    \includegraphics[scale=0.27]{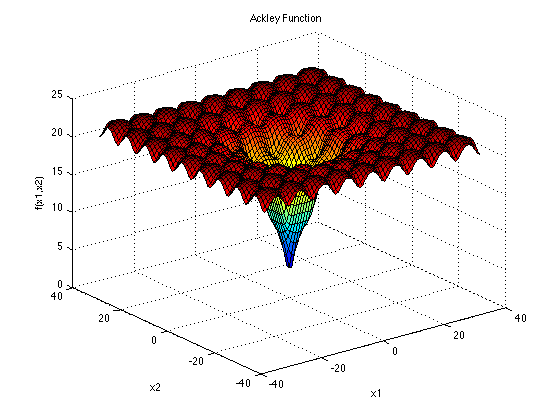}
    \subcaption{Ackley}
     \label{fig:rosenbrock}
\end{subfigure}
\caption{Three benchmark functions: Rosenbrock, Rastrigin and Ackley.}
\label{fig:functions}
\end{figure*}

We use popular mathematical benchmark functions from the field of global optimization to train our NN-proxies. Our choice of functions are Rosenbrock, Rastrigin and Ackley. Figure \ref{fig:functions} shows the functions plots. 

\noindent \textbf{Rosenbrock} \cite{rosen} is usually evaluated on the hypercube $x_i \in [-5, 10], \enskip  \forall \enskip i = 1, …, d$. We have restricted to the hypercube $x_i \in [-2.048, 2.048], \enskip  \forall \enskip i = 1, …, d$ and use it in our study for the sake of simplification. The mathematical formulation is stated in Eq. \ref{eq:rosenbrock}. The global minimum for this function is at $\mathbf{x} = [1,..,1]$ and $f(x_i)=0$.

\noindent \textbf{Rastrigin} \cite{rastrigin1974systems} is usually evaluated on the hypercube $x_i \in [-5.12, 5.12], \enskip \forall \enskip i = 1, …, d$, and the mathematical formulation for this function is stated in Eq.  \ref{eq:rastrigin}). The global minimum is at $\mathbf{x} = [0,..,0]$ with $f(x_i)=0$.

\noindent \textbf{Ackley} \cite{d.h.ackley1987a-connectionist} is usually evaluated on the hypercube $x_i \in [-32.768, 32.768] \enskip \forall \enskip i = 1, …, d$, it may be restricted to a smaller domain. For the mathematical representation of this function, see Eq. \ref{eq:ackley}. The global minimum is at $x_i = [0, ... , 0]$ with  $f(x_i) = 0$.

\begin{equation}\label{eq:rosenbrock}
    f(x) = \sum_{i=1}^{d-1}[100(x_{i+1} - x_{i}^{2}) ^ {2} + (x_i - 1)^2] 
\end{equation}

\begin{equation}\label{eq:rastrigin}
   f(x) = 10d + \sum_{i=1}^{d}[x_{i}^{2} - 10 cos(2\pi x_i)] 
\end{equation}

\begin{equation}\label{eq:ackley}
    f(x) = -20 exp(-0.2 \sqrt{\frac{1}{d} \sum_{i=1}^d x_i^2}) - exp(\frac{1}{d} \sum_{i=1}^d cos(2\pi x_i)) + 20 + e 
\end{equation}

\section{Optimization Methods}
\label{sec-app-optmethods}
\subsection{Particle Swarm Optimization}

Particle Swarm Optimization was proposed by Kennedy and Eberhart \cite{PSO}. It is inspired by the idea that a school of fish or a flock of birds that moves in a group `can profit from the experience of all other members'. A brief overview of the underlying mechanism is described below.

\noindent \textbf{Algorithm Details.} Let there be $P$ particles, where we denote the position of particle $i$ at iteration $t$ as $X^i(t)$. This position can then be stated as $X^i(t) = (x^{i}(t),y^{i}(t))$ and velocity as $V^i(t)=(v_x^i(t), v_y^i(t))$. At the next iteration, the position is updated as: $X^i(t+1) = X^i(t)+V^i(t+1)$ and the velocity is updated as: 
\begin{equation}
    V^i(t+1) = w V^i(t) + c_1r_1(pbest^i – X^i(t)) + c_2r_2(gbest – X^i(t))
\end{equation}
Where $r_1$ and $r_2$ are random numbers between 0 and 1. Constants $w$, $c_1$ and $c_2$ are inertia, learning rate for individual ability and social influence respectively. $pbest^i$ is the position that gives the best cost explored by particle $i$ and $gbest$ is the best explored solution by all swarms. 

\subsection{Genetic Algorithm}

The genetic algorithm \cite{GA} is a search heuristic that is inspired by Darwin’s theory of natural evolution in which the fittest of individuals are the ones who survive. This fitness is measured by a fitness function. At every iteration, individuals with high fitness have more chance to be selected for reproduction to produce off springs for the next generation. The algorithm terminates if the population has converged i.e, offspring are not different from the previous generation or number of generations has been reached. For details related to implementation, see for example \cite{deb2002ga}.

\section{Experiments: Additional Details}
\label{sec-app-exp}
In this section we provide details about our experimental settings including the proxy models' architecture and training details. 

We employ three different sampling mechanisms. For the dense sampling strategy, we use 10,000 samples for all cases to build the NN-proxies and these samples are equally spaced in the hyperspace. For sparse case, we use 25\% of the data that we use in the dense case. For the case of Gaussian sampling, we use the same number of points as the dense case, but the distribution follows a multidimensional Gaussian distribution.

In Dense case we use 10K points for training our models. 
In Sparse case we use 25\% of the data. 
In Gaussian case, we use 10K centred around the functions' global minimum.

For Rosenbrock proxy model in 2D case, we employ a 3 layer multilayer perceptron (MLP) with hidden units of $\{15, 50, 15\}$ and RelU activation function. We train for 100 epochs with learning rate of 0.001 and Adam optimizer. In 4D case, we use a 4 layer MLP with hidden units of $\{15,50,15,10\}$ and RelU activation. We train with the same number of epochs and similar learning rate and optimizer as 2D case. In 10D case, we use the same model setting as 4D case and train for 500 epochs with learning rate of 0.001 and Adam optimizer. All experiments were run on a NVIDIA RTX 6000 graphics card.

For Rastrigin and Ackley proxy models in 2D, 4D and 10D cases, we use 6 layer MLP with hidden units of $\{20,50,120,70,20,10\}$. We use Relu activation and train for 500 epochs. Adam optimizer is used with learning rate of 0.001.

\end{document}